%% file: elsarticle-template-harv.tex
\documentclass[preprint,12pt]{elsarticle}

\usepackage{amssymb,amsmath}
\usepackage{epsfig}
\usepackage[utf8]{inputenc}
\usepackage{multirow}
\usepackage{amssymb} 
\usepackage{graphicx}
\usepackage[usenames,dvipsnames,svgnames,table]{xcolor}
\usepackage{colortbl}
\usepackage{caption}
\usepackage{subcaption}
\usepackage{setspace}
\usepackage[linesnumbered,ruled,vlined]{algorithm2e}
\usepackage{scalefnt} %tamanho de tabelas
\usepackage{pdflscape} %modo paisagem
\usepackage{afterpage}
\usepackage{rotating}
\usepackage{tabularx,booktabs}
\usepackage{makecell}
\usepackage{listings}
\usepackage{multirow}
\usepackage{dblfloatfix}
\usepackage{soul} 
\usepackage{etoolbox}
\usepackage{todonotes}
\usepackage{bold-extra}
\usepackage{contour}
\usepackage{wasysym}
\usepackage{tablefootnote}

\contourlength{0.01em}

\newcommand{\vspacedef}{\vspace{-0.0em}}

\usepackage{setspace}

\definecolor{mygreen}{rgb}{0.0, 0.6, 0.0}

\journal{CORR}

\begin{document}

\newdefinition{definition}{Definition}

\begin{frontmatter}



\cortext[cor1]{Main corresponding author.}

\title{Discovering Heterogeneous Subsequences for Trajectory Classification}

 \author[ifsc]{Carlos Andres Ferrero}
 \ead{andres.ferrero@ifsc.edu.br}
 \address[ifsc]{Federal Institute of Santa Catarina Lages, Santa Catarina, Brazil}
 \author[ufsc]{\\Lucas May Petry} 
 \author[ufsc]{Luis Otavio Alvares}
 \author[unila]{Willian Zalewski}
 \address[unila]{Federal University for Latin American Integration, Foz do Iguassu, Parana, Brazil}
 \author[ufsc]{Vania Bogorny}
 \address[ufsc]{Universidade Federal de Santa Catarina, UFSC, Florianópolis, Santa Catarina, Brazil}


\end{frontmatter} 


\input{1_Introduction.tex}

\input{3_Proposal.tex}

\input{4_Experiments.tex}

\input{5_Conclusions.tex}

%\section*{Acknowledges}

%\section*{References}
\small
\setlength{\bibsep}{0.3em}
\bibliographystyle{plain} 
%\bibliography{references}

\end{document}

%% file: 1_Introduction.tex
\section{Introduction}
\label{sec:introduction}

Trajectory classification is an important issue in mobility data mining, since it is used for several discovering transportation modes, animal categories, hurricane strengths 

%For many years semantic trajectory analysis has been performed over a single point of view. Example...

There are different works in the literature about semantic trajectory classification, mainly focusing on the next place prediction, such as the place category or the geographic location~\cite{cho2011friendship}. However these works are limited to only consider three dimension, as space, time, and semantics. Ferrero in \cite{Ferrero2016} introduced the concept of Multiple Aspect Trajectory Analysis, that consists of analyzing trajectory data by integrating other movement aspects to further enrich trajectory data, such as more information about the visited places, the transportation modes, the weather conditions, and the social interactions.

\begin{figure}[!htpb]
\centering
\includegraphics[width=0.8\textwidth]{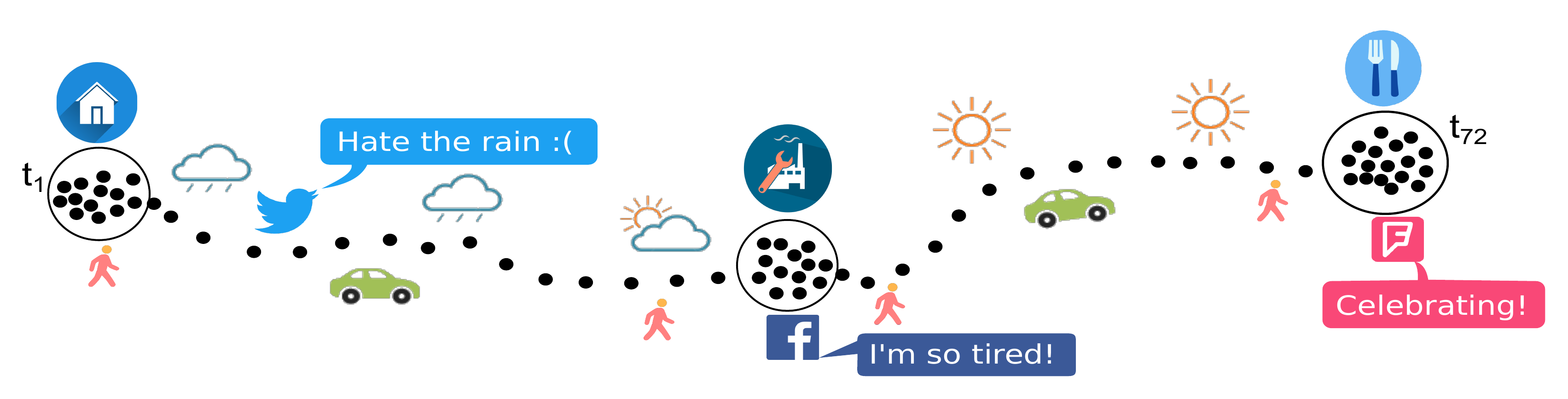}
\caption{An example of Multiple Aspect Trajectory.}
\label{fig:mat}
\end{figure}

The proposal in~\cite{Ferrero2016} is that time has come to integrate all relevant information about movement in trajectories and explore trajectory analysis over several layers of information. An example of this new kind of trajectory is shown in Figure~\ref{fig:mat}. In this figure, an individual starts his/her trajectory at home, then he/she goes to work by car, and after work, he/she goes to eating at a restaurant. Note that during his/her movement the weather condition changes two times, from rainy to cloudy and then to sunny, and part of the trajectory is carried out on foot and part by car. In addition, the individual uses different social networks (e.g. Twitter, Facebook, and Foursquare) to post how he/she is feeling.

Multiple aspect trajectory analysis and mining is the research topic of a current European Research Project 
H2020-MSCA-RISE-2017 called \textsc{MASTER} (Multiple ASpect TrajectoriEs Representation and analysis).

This new representation of multiple aspect trajectories has the challenge of dealing with multiple and heterogeneous dimensions. Because this is a recent and new kind of data, there are no works in the literature to analyze and to extract new knowledge about user' behavior, which can help to improve location services, privacy measures, safety strategies, and others.

%An interesting recent 
An emerging classification task in trajectory data is to learn discriminant parts of trajectories, called subtrajectories, that characterize the behavior of an individual or a group of individuals. We claim that discovering discriminant subtrajectories from this new kind of data is very important to several supervised tasks, such as building classification models to predict who is the user of a trajectory based on its movement and extracting the movement profile to describe the most important movements that characterize itself. These tasks can play an essential role in privacy protection applications.

\begin{figure}[!htbp]
\centering
\vspace{-0.5em}
\includegraphics[width=1.0\textwidth]{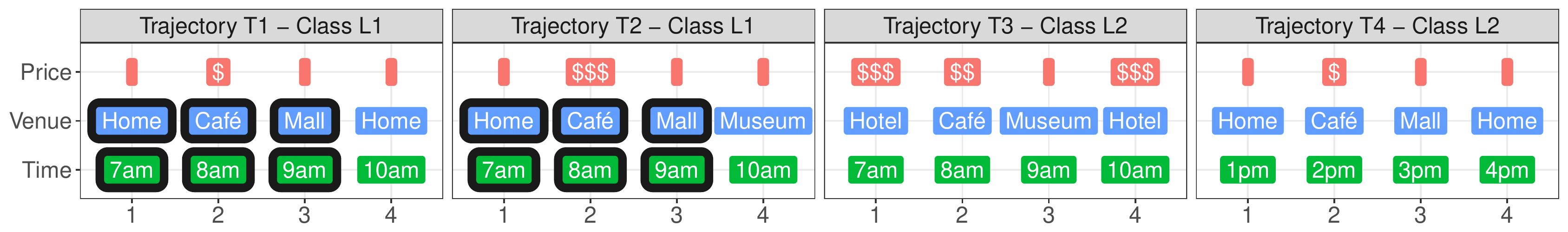}
\caption{The problem of finding relevant subtrajectories in multidimensional trajectories.}
% Dimensoes 3.5 x 13
\label{fig:multidimensionalProblem}
\end{figure}

In previous work, Ferrero in~\cite{ferrero2018} proposed a novel approach for learning discriminant subtrajectories, called \textsc{Movelets}. The method was inspired on time series \emph{shapelets}~\cite{Ye2011}. \textsc{Movelets} is a parameter-free method and supports multiple dimensions. However, all dimensions need to be considered together, what significantly reduces the efficiency of the learning process, because it is more difficult to find trajectory patterns as the number of dimensions increases. In addition, by analyzing each dimension individually limits the mining task, because the interaction among dimensions is a key issue in multiple aspect trajectory data analysis. For instance, Figure~\ref{fig:multidimensionalProblem} shows four trajectories $T_1$ to $T_4$ of two class users $L_1$ and $L_2$, that visit \textit{cafés}, \textit{hotels}, \textit{museums}, \textit{shops}, with different prices. These trajectories are sequences of Foursquare check-ins represented by dimensions \textit{Time}, \textit{Venue} and venue's \textit{Price}. Suppose that we want to find a subtrajectory of three check-ins that discriminate the user $L_1$ from the user $L_2$. Figure~\ref{fig:multidimensionalProblem} highlights (thick line around check-ins) a discriminant subtrajectory in trajectories $T_1$ and $T_2$ of class $L_1$ that not happens in trajectories $T_3$ and $T_4$ of class $L_2$. 

\vspace{-0.5em}
Note in Figure~\ref{fig:multidimensionalProblem} that by considering all dimensions together there are no subtrajectories of three sequential check-ins in common  between $T_1$ and $T_2$, because $T_1$ goes in a café of price \$ while $T_2$ goes in café of price \$\$\$. By analyzing each dimension individually, the dimension \textit{Time} is not discriminative, because $T_1$ has check-ins at the same time of $T_3$, at 7am, 8am, 9am, and 10am. Also the dimension \textit{Venue} is not discriminative, because $T_1$ and $T_4$ have both check-ins in \textit{Home, Café, Mall, and Home}. The problem is that by analyzing all dimensions together or all dimensions individually is not possible to find a discriminant subtrajectory of three sequential check-ins. So, \textsc{Movelets} is not useful for this task.

To find the discriminant subtrajectory \textit{highlighted} in Figure~\ref{fig:multidimensionalProblem} it is necessary to explore dimension combination as well as the sequence of check-ins. There are no methods in the literature to solve the problem of finding relevant subtrajectories in trajectory data represented by multiple dimensions.

In this work, we propose a new method for discovering relevant subtrajectories in multidimensional and heterogeneous sequences, focusing on movement trajectories. Our method explores and finds the dimension combination to achieve the highest discriminative power. The main contributions of our work are summarized as follows:

\begin{enumerate}

\item A new method, called \textsc{MasterMovelets}, for discovering relevant subtrajectories that considers multiple and heterogeneous dimensions. The method is parameter-free and domain independent. %This method can be used in several applications on trajectory data, such as classification, knowledge extraction, anonymization, among others.

\item A new method for multidimensional alignment based on ranking, called \textsc{MasterAlignment}. The method discovers the part of a trajectory that is most similar to a subtrajectory considering multiple dimensions together.

\item A new method to measure the subtrajectory relevance over multiple and heterogeneous dimensions, called \textsc{MasterRelevance}.

\end{enumerate}

The remaining of this paper is structured as follows:   Section~\ref{sec:proposal} describes the proposed method, Section~\ref{sec:experiments} shows the experiments, and Section~\ref{sec:conclusions} conclusions and future work.

%% file: 3_Proposal.tex
\vspace{-0.5em}
\begin{sloppypar}
\section{Discovering Relevant Multiple Aspect Subtrajectories}
\label{sec:proposal}
\end{sloppypar}

In this section we detail our proposal for finding relevant subtrajectories in trajectories represented by multiple and heterogeneous dimensions. Section~\ref{sec:basicDefinitions} introduces basic definitions, Section~\ref{sec:masterMovelets} describes the method for discovering multidimensional \emph{movelets}, Section~\ref{sec:multidimensionalAlignment} presents a new method for multidimensional subtrajectory alignment, and Section~\ref{sec:MQM} presents a new method for measuring the quality of multidimensional subtrajectories.

\vspace{0.3em}
\subsection{Basic Definitions}
\label{sec:basicDefinitions}
%\begin{definition}
%\label{def:trajectory}.
%\textbf{Trajectory}. 
%A \emph{trajectory} $T = \langle p_1, ..., p_m \rangle$ is a sequence of time-ordered  points $p_i=(x,y,t)$, where $x,y$  correspond to the spatial location of the object at the time instant $t$.
%\end{definition}

%A \emph{multidimensional and heterogeneous trajectory} $T = \langle p_1, ..., p_m \rangle$ is a sequence of time-ordered movement information $p_i=(c_1, c_2, c_3, ..., c_d)$ of a moving object, where $c_1$ and $c_2$ are spatial and temporal dimensions, respectively, an the order dimensions $c_3$ to $c_d$ represent additional information about the moving object at this point, such as the information about the place that it visits, the weather condition, and others.

A multidimensional trajectory $T$ is a sequence of elements $\langle e_1, e_2, \dots, e_m \rangle$, where each element has a set of $l$ dimensions $D=\{d_1, d_2, \dots, d_l\}$. Our goal is to find parts of multidimensional trajectories with high discriminative power among classes. A part of a trajectory is called \emph{subtrajectory}. So given a trajectory $T$ of length $m$, a \emph{subtrajectory} $s = \langle e_a, \ldots, e_b \rangle$ is a contiguous subsequence of $T$ starting at point $e_a$ and ending at point $e_b$, where $1 \leq a < m$ and $a \leq b \leq m$. The subtrajectory $s$ can be represented by all dimensions $D$ or a subset of dimensions $D' \subseteq D$. The length of the subtrajectory is defined as $w = |s|$. In addition, we also define the set of all \emph{subtrajectories} of length $w$ in $T$ as $S_{T}^{w}$, and the set of all \emph{subtrajectories} of all lengths in $T$ as $S_{T}^{*}$.

In order to find discriminant parts of a trajectory we need to define the distance between two subtrajectories. This distance may consider the dimensions of the problem, since a point may have multiple and heterogeneous dimensions, we formally define the concept of \emph{distance between elements}.

\begin{sloppypar}
\vspacedef
\begin{definition}
\label{def:dist_two_points}
\textbf{Distance vector between two multidimensional elements}.
Given two elements $e_i$ and $e_j$ represented by $D$ dimensions, the distance between two multidimensional elements $dist(e_i,e_j)$ returns a \emph{distance vector} $V=(v_1, v_2, ..., v_{d})$, where each $v_k = dist\_e_k(e_i,e_j)$ is the distance between two elements at dimension $k$, that respects the property of symmetry $dist\_e_k(e_i,e_j) = dist\_e_k(e_j,e_i)$.
\end{definition}
\vspacedef
\end{sloppypar}

The idea behind Definition~\ref{def:dist_two_points} is to allow of using a distance function for each dimension and storing them into a \emph{distance vector} to help computing the distance between two subtrajectories of equal length, which is given in Definition~\ref{def:dist_two_subtrajectories_eq}.

%Definition~\ref{def:dist_two_points} allows the definition of several distance measures between trajectory points by combining distance functions for each dimension (such as position, time, speed,  acceleration, direction, and so on). In addition, it can be used to compute the distance between two subtrajectories of equal length, as in Definition~\ref{def:dist_two_subtrajectories_eq}.

\begin{sloppypar}
\vspacedef
\begin{definition}
\label{def:dist_two_subtrajectories_eq}
\textbf{Distance vector between two subtrajectories of equal length}.
Given two subtrajectories $s$ and $r$ both of length $w$ and dimensions $D$, $dist\_s(s,r)$ computes the pairwise distance between their sequential elements $(V_1, V_2, \ldots, V_w)$ in a \emph{distance vector} $\mathbf{V}=(\mathbf{v}_1, \mathbf{v}_2, \ldots, \mathbf{v}_d)$, where each $\mathbf{v}_k$ is the distance value between $s$ and $r$ at dimension $k$, which is obtained by a function over the $w$ distances between the two subtrajectories at dimension $k$. Each distance $\mathbf{v}_k$ respects the property of symmetry $dist\_s_k(s_k,r_k) = dist\_s_k(r_k,s_k)$.
\end{definition}
\vspacedef
\end{sloppypar}

To evaluate whether a subtrajectory $s$ is into a trajectory $T$, we need to find the most similar \emph{subtrajectory} of $T$ to the \emph{subtrajectory} $s$. The most similar subtrajectory of $T$ to $s$ is called \emph{best alignment}, and is a subtrajectory $r$ with the minimum distance. This comparison is given in Definition~\ref{def:dist_best_match_traj}.

\begin{sloppypar}
\vspacedef
\begin{definition}
\label{def:dist_best_match_traj}
\textbf{Distance vector between trajectory and subtrajectory.}
Given a trajectory $T$ and a subtrajectory $s$ of length $w = |s|$, the distance between them is the best alignment of $s$ into $T$, which is defined by $W^T_s = min( dist\_s(s,r)\;|\;r \in S_{T}^{w})$, where $S_{T}^{w}$ is the set of all subtrajectories of length $w$ into $T$, and $min()$ returns the smallest \emph{distance vector} that is the best alignment between $s$ and all subtrajectories in $S_{T}^{w}$.
\end{definition}
\vspacedef
\end{sloppypar}

The number of all possible subtrajectories of any length in a trajectory problem with $n$ trajectories of length at most $m$, and dimensions $d$, is $O(n \times m^2 \times 2^d)$. By representing trajectories using all subtrajectories as features, the induction of classification models is impracticable, because of the relation between instances and attributes. So, the selection of only the most \emph{relevant subtrajectories} is necessary. Before defining a relevant subtrajectory we define a subtrajectory candidate in Definition~\ref{def:subtrajectory_candidate}.

\begin{sloppypar}
\vspacedef
\begin{definition}
\label{def:subtrajectory_candidate}
\textbf{Subtrajectory Candidate}. 
A \emph{subtrajectory candidate} is a tuple $(T,start,end,C,\mathbb{W},L)$, where $T$ is the trajectory that origins the candidate; $start$ and $end$ are the positions that the candidate begins and ends, respectively; $C$ contains the candidate dimensions; $\mathbb{W}$ is a set of pairs $(W_i,class_i)$, where $W_i$ is the distance vector of the best alignment into a trajectory $T_i$ and $class$ is the class label of $T_i$; $L$ is a pair $(sp,score)$, where $sp$ contains the \emph{split points} for each dimension and $score$ is the relevance score.
%\vspaceitem
%\vspace{-1em}
% \begin{itemize}
%  \itemsep0em
%  \item $T$ is the trajectory that origins the candidate;
%  \item $start$ is the position in $T$ where the candidate begins;
%  \item $end$ is the position in $T$ where the candidate ends;
%  \item $C$ is the set of dimension combination;
%  \item $\mathbb{D}$ is a set of pairs $D=(V,class)$, where $V$ is the distance vector to the best alignment into a trajectory $T'$ (Definition~\ref{def:dist_best_match_traj}) and $class$ is the class label of $T'$; 
%  \item $Q$ is pair $(sp,score)$, where $sp$ contains the \emph{split points} for dimensions and $score$ is the relevance score.
%  \end{itemize}
\end{definition}
%\vspacedef
\end{sloppypar}

Evaluating the \emph{relevance} of each subtrajectory is fundamental to explore \emph{movelets}. In classification problems this relevance is given by the capability to differentiate trajectories of one class (\emph{target class}) from trajectories of other classes. In other words it is expected that a relevant subtrajectory appears into trajectories of the \emph{target class} and does not appear into trajectories of other classes. To measure the relevance we use the set $\mathbb{W}$ to find a set of distance split points $sp$ in order to split the set of distances into two subsets: the \emph{left side} with the nearest distances, from $0$ to $sp$; and the \emph{right side} with the  longest distances, from $sp$ to $\infty$. The former contains only distances to trajectories of the \emph{target class} and the latter the other distances. The greater the number of distances in the \emph{left side} the greater the relevance of the subtrajectory candidate. %So the subtrajectory relevance is related to the capability to find \emph{split points} that better separate the classes. 
Based on the concept of \emph{relevance} we define a \emph{movelet} as in given in Definition~\ref{def:movelet}.

\begin{sloppypar}
\vspacedef
\begin{definition}
\label{def:movelet}
\textbf{Movelet}. 
Given a trajectory $T$ and a subtrajectory candidate $s \in T$, the subtrajectory $s$ is a \emph{movelet} if for each subtrajectory $r \in T$ that overlaps $s$ in at least one element, $s.L.score > r.L.score$. 
\end{definition}
\vspacedef
\end{sloppypar}

In other words, a subtrajectory is considered as a \emph{movelet} if there is no other candidate overlapping it with more relevance on any dimension combination. The \emph{movelet} discovery consists of exploring all subtrajectory candidates from a trajectory training set and selecting only the subtrajectories with highest relevance, which are called \emph{movelets}. 

\subsection{\textsc{MasterMovelets}: Multidimensional Movelets Discovering Algorithm}
\label{sec:masterMovelets}

\begin{sloppypar}
In this section we present the algorithm for discovering multidimensional \emph{movelets}, called \textsc{MasterMovelets} (Multiple ASpect TrajEctoRy Movelets). This method consist of an extension of the algorithm to extract classical \textsc{Movelets} with support to multiple and heterogeneous dimensions. This method is detailed in Algorithm~\ref{alg:best_movelets}, that has as the unique input the trajectory training set $\mathbf{T}$, without any parameter. The output is the set of \emph{movelets}. 
\end{sloppypar}

\begin{algorithm}[!h]
  \footnotesize
  \caption{\textsc{MasterMovelets}}
  \label{alg:best_movelets}
  \SetAlgoLined
  \SetKwInOut{Input}{Input}
  \SetKwInOut{Output}{Output}
  \Input{$\mathbf{T}$ // trajectory training set}	
  \Output{$movelets$  // set of relevant subtrajectories}
  $movelets \leftarrow \emptyset$ \;  
  \For{\textbf{each} trajectory T \textbf{in} $\mathbf{T}$\label{alg1:forTrajectoryStart}}{	
    	$candidates \leftarrow \emptyset$\;
		%$m_T \leftarrow T.length$\;
		$A_{0} \leftarrow ComputeElementDistances~(T,\mathbf{T})$\label{alg1:cpd} \;   
        \For{subtrajectory length $w$ \textbf{from} $1$ \textbf{to} $T.length$ \label{alg1:forLengthStart}}{
        \textbf{$A_{w} \leftarrow CSD~(A_{w-1}, A_{0}, w)$}\label{alg1:csd} \;
        %$candidates \leftarrow \emptyset$\;
          \For{$position$ $j$ \textbf{from} $1$ \textbf{to} $(T.length-w+1)$ \label{alg1:forEachPositionStart}}{          
         % STEP 1 - PRECALCULATE THE DISTANCE RANKING FOR EACH TRAJECTORY AND EACH DIMENSION
         % It takes O(n x d x m log m)
          $R \leftarrow \emptyset$\label{alg:DBDC:Rempty} \;

		   \For{trajectory $i$ \textbf{from} $1$ \textbf{to} $|\textbf{T}|$\label{alg:DBDC:forTrajectory1Start}}{

          \For{dimension $d$ \textbf{from} $1$ \textbf{to} $|D|$ \label{alg2:forRankStart}}{

	          $R[i,d,..] \leftarrow Rank(A_w[i,j,d,..])$ \;

        	}\label{alg2:forRankEnd}

		}\label{alg:DBDC:forTrajectory1End}		
        $bestScore \leftarrow 0$ \;
        \For{\textbf{each} $dimension~combination$ $C$ \textbf{in} $C^*_d$ \label{alg1:forEachDimensionCombinationStart}}{

		$\mathbb{W} \leftarrow \emptyset$\label{alg:DBDC:Dempty}\;
		% It takes O(n * m * d)
		\For{trajectory $i$ \textbf{from} $1$ \textbf{to} $|T|$ \label{alg:DBDC:internalForStart}}{
        		
			\underline{$W_i \leftarrow \mathit{min}$~\textsc{MasterAlignment}$(R[i,C,..], A_w[i,j,C,..])$\label{alg:DBDC:bestAlignment} \;}

			$\mathbb{W}[i] \leftarrow (W_i, \mathbf{T}[i].class)$ \label{alg:DBDC:storingdistance} \;     
     	}\label{alg:DBDC:internalForEnd}

		% It takes O(n^2)
      	\underline{$relevance \leftarrow \mathit{assess}$~\textsc{MasterRelevance}$(\mathbb{W},T.class)$}
        \label{alg:DBDC:AssessQuality} \;
        
        % It takes O(n)
      	\If{$relevance.score > bestScore$\label{alg:DBDC:minBestStart}}{
        	$bestL \leftarrow relevance$ \;
            $best\mathbb{W} \leftarrow \mathbb{W}$ \;
            $bestC \leftarrow C$ \;
            $bestScore \leftarrow relevance.score$ \;      		  	
      	}\label{alg:DBDC:minBestEnd}

            }\label{alg1:forEachDimensionCombinationEnd}
          $s \leftarrow SubtrajectoryCandidate\big(T, j, (j + w - 1), bestC, best\mathbb{W}, bestL\big)$\label{alg:DBDC:defineCandidate} \;
          $candidates \leftarrow candidates \cup s$\label{alg:DBDC:addToCandidates} \;
        }\label{alg1:forEachPositionEnd}	      
          $trajectoryCandidates \leftarrow trajectoryCandidates \cup candidates$\label{alg1:addToTrajectoryCandidates}\;          
        }\label{alg1:forLengthEnd}        
        $SortByQuality~(trajectoryCandidates)$\label{alg1:SortByQualityE}\;      	  
    	$RemoveSelfSimilar~(trajectoryCandidates)$\label{alg1:RSSE}\;
        $movelets \leftarrow movelets \cup trajectoryCandidates$\label{alg1:union1}\;
    }\label{alg1:forTrajectoryEnd}
   \Return{$movelets$}
\end{algorithm}

Algorithm~\ref{alg:best_movelets} finds for each trajectory in the training set $\mathbf{T}$ the most relevant subtrajectories considering multiple dimensions (lines~\ref{alg1:forTrajectoryStart} to \ref{alg1:forTrajectoryEnd}). For each $T \in \mathbf{T}$ it computes all distances between all trajectory elements in $T$ and all elements in $\mathbf{T}$, and stores them into the 
4-dimensional array $A_0$ of distances. Each value $A_0[i,j,d,k]$ the distance between the element of $T$ at position $j$ and the element of $T_i \in \mathbf{T}$ at position $k$, considering dimension $d$. The distance array is precomputed in order to perform this computation only once (line~\ref{alg1:cpd}). Next, it explores all subtrajectory lengths, one by one (lines~\ref{alg1:forLengthStart} to \ref{alg1:forLengthEnd}). For a length $w$ it computes the array of distances $A_w$, using the distance values computed for subtrajectories of length $(w-1)$ and for the elements, represented by $A_{w-1}$ and $A_0$, respectively (line~\ref{alg1:csd}). $A_w$ contains all the distance sums at subtrajectory length $w$. Next, for each subtrajectory in $T$ of size $w$ starting at the $j$th position, it discovers the best dimension combination for the subtrajectory based on $A_w$, and adds it into the $candidates$ set (lines~\ref{alg1:forEachPositionStart}-\ref{alg1:forEachPositionEnd}). In this loop the algorithm first computes for each subtrajectory in $T$ starting at position $j$ the distance ranking $R$ among all starting positions in the $i$th trajectory, at dimension $k$ (lines~\ref{alg:DBDC:Rempty} to~\ref{alg:DBDC:forTrajectory1End}). Then, for each dimension combination $C$ of all dimension combination $C^*_d$ it computes the best alignment between the subtrajectory in $T$ starting at position $j$ to each trajectory $T_i$, using a specific method for multidimensional alignment, called \textsc{MasterAlignment} , storing the distance vector into $\mathbb{W}$ (lines~\ref{alg:DBDC:Dempty} to~\ref{alg:DBDC:internalForEnd}). After that, the algorithm evaluates the set of distances in $\mathbb{W}$ using a specific function for measuring the relevance and discovering the split points, called \textsc{MasterRelevance}, and preserves the relevance and the distances of the best dimension combination (lines~\ref{alg:DBDC:AssessQuality} to~\ref{alg:DBDC:minBestEnd}). Then, it defines the subtrajectory candidate as the subtrajectory with the most relevant dimension combination and stores it into the set $candidates$ (lines~\ref{alg:DBDC:defineCandidate} and~\ref{alg:DBDC:addToCandidates}). Next, it stores the subtrajectory candidates of any length into $trajectoryCandidates$ (line~\ref{alg1:addToTrajectoryCandidates}). Following the external loop, it sorts the trajectory candidates by their relevance and removes those \textit{self similar} (lines~\ref{alg1:SortByQualityE} to \ref{alg1:RSSE}). Two candidates are \textit{self similar} if they are overlapping on at least one point and the algorithm preserves the highest relevance candidate. Finally, it adds the remaining candidates to the $movelets$ set. 

Two key points to perform \emph{movelets} discovery in trajectories represented by \emph{multiple and heterogeneous dimensions} are: finding the best alignment of the subtrajectory into a trajectory, performed by the method \textsc{MasterAlignment} and measuring the relevance of subtrajectories, performed by the method \textsc{MasterRelevance}. These key points substantially change the way to discover classical \emph{movelets} and are detailed in the next sections.

\subsection{Multidimensional Alignment of a Subtrajectory into a Trajectory}
\label{sec:multidimensionalAlignment}

The best alignment between a subtrajectory and a trajectory consist of finding the most similar part of the trajectory to the subtrajectory. The function $min()$ in Definition~\ref{def:dist_best_match_traj} performs the best alignment and returns the distance. In the case of one-dimensional alignment ($|D|=1$) the function returns only the minimum distance value, but in the case of $|D|>1$ all distance values of the $D$ dimensions must be considered, in the form of a distance vector. An ingenuous solution consists of transforming each \emph{distance vector} in a unique value by normalizing and weighting the dimensions, but this brings two major drawbacks. The first is that the normalization requires to previously know some information about the distance distribution, such as the mean, standard deviation, or maximum and minimum values, which is totally domain dependent. The second is that the predefinition of dimension weights for all classes exclude the possibility of capturing the particular relevant dimensions for each class.

%however this brings disadvantages in the presence of heterogeneous dimensions, because of the distance scales are different and domain dependent, and the weights of each dimension are unknown or is not possible to define static values.

%To exemplify, Figure~\ref{fig:bestAlignmentA} shows an example of a subtrajectory and Figure~\ref{fig:bestAlignmentB} a trajectory. In the trajectory of Figure~\ref{fig:bestAlignmentB} the object: ``\textit{at 7am, stays at an expensive (\$\$\$) hotel; then walks to the park at 07:15am; has a cheap (\$) breakfast at 8am at a Café; arrives at work 8:30am, and after an hour goes to an expensive (\$\$\$) Mall to meet his/her boss. They use to drink coffee in an very expensive (\$\$\$\$) Café for talking business. Finally, the object returns to work at 11:30am.}''. The subtrajectory we want to align is: ``\textit{Users that visit an expensive (\$\$\$) Café around 10:30am and after go to work around 11:00am}''.

\begin{figure*}[!hptb]
\centering

\begin{subfigure}[t]{0.30\textwidth}
\centering\footnotesize
\caption{Subtrajectory $s$}
\vspace{-0.5em}
\includegraphics[height=8em]{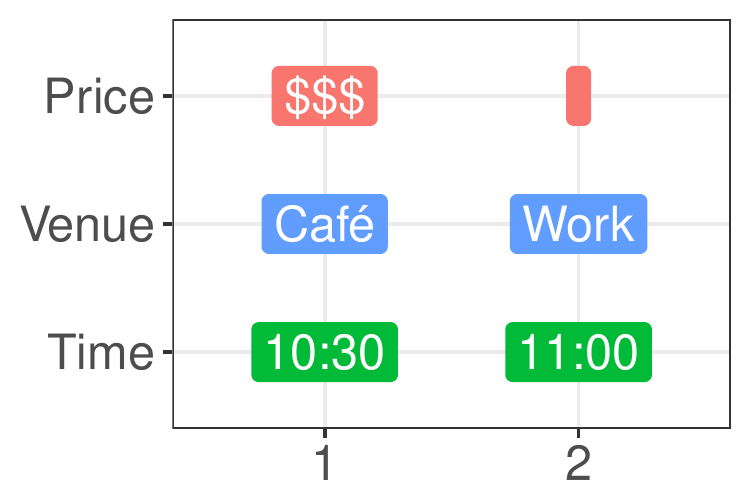}
\label{fig:bestAlignmentA}
\end{subfigure}
~
\begin{subfigure}[t]{0.65\textwidth}
\centering\footnotesize
\caption{Trajectory $T$}
\vspace{-0.5em}
\includegraphics[height=8em]{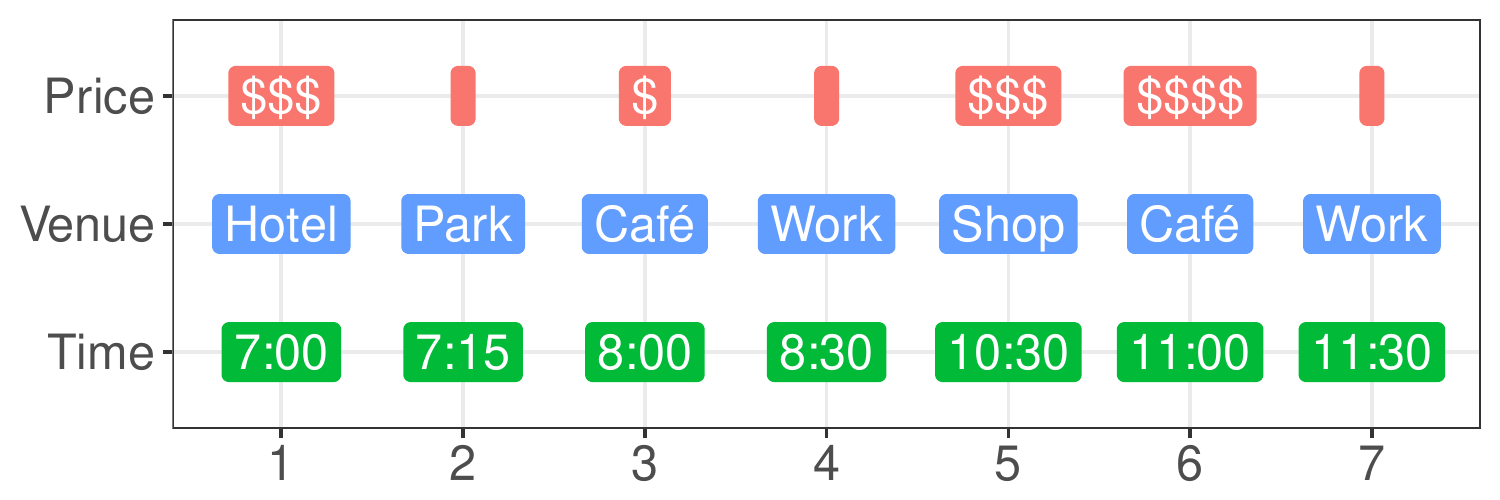}
\label{fig:bestAlignmentB}
\end{subfigure}
\vspace{-1.5em}
\caption{Example of a subtrajectory $s$ and a trajectory $T$.}
\label{fig:bestAlignment}
\end{figure*}

To exemplify, Figure~\ref{fig:bestAlignmentA} shows an example of a subtrajectory $s$ and Figure~\ref{fig:bestAlignmentB} a trajectory $T$. In the trajectory of Figure~\ref{fig:bestAlignmentB} the object: ``\textit{at 7am, stays at an a hotel of price \$\$\$; then walks to the park at 07:15am; has breakfast at 8am at a Café of price \$; arrives at work 8:30am, and after an hour goes to a Shop to buy a gift to his/her boss. They use to drink coffee in a Café of price \$\$\$\$ for talking business. Finally, the object returns to work at 11:30am.}''. The subtrajectory we want to align is: ``\textit{Users that visit a Café of price \$\$\$ around 10:30am and after go to work around 11:00am}''.

Based on Figure~\ref{fig:bestAlignment} we consider the following insights about alignments:

\begin{enumerate}
\item For the dimension \emph{Time}, the best alignment of the subtrajectory of Figure~\ref{fig:bestAlignmentA} into the trajectory of Figure~\ref{fig:bestAlignmentB} happens between the check-ins 5 and 6. But note that the venues are different (\textit{Café} $\neq$ \textit{Shop} and \textit{Work} $\neq$ \textit{Café}).

\item For the dimension \emph{Venue}, there are two best alignments: the first between the check-ins 3 and 4 and the other between the 6 and 7.

\item On the \emph{Price} dimension, the best alignment happens between the check-ins 1 and 2, where the venues are also different. %(\textit{Café} $\neq$ \textit{Hotel} and \textit{Work} $\neq$ \textit{Park}).
\end{enumerate}

\begin{figure}[!htbp]
\centering\footnotesize
\includegraphics[height=8em]{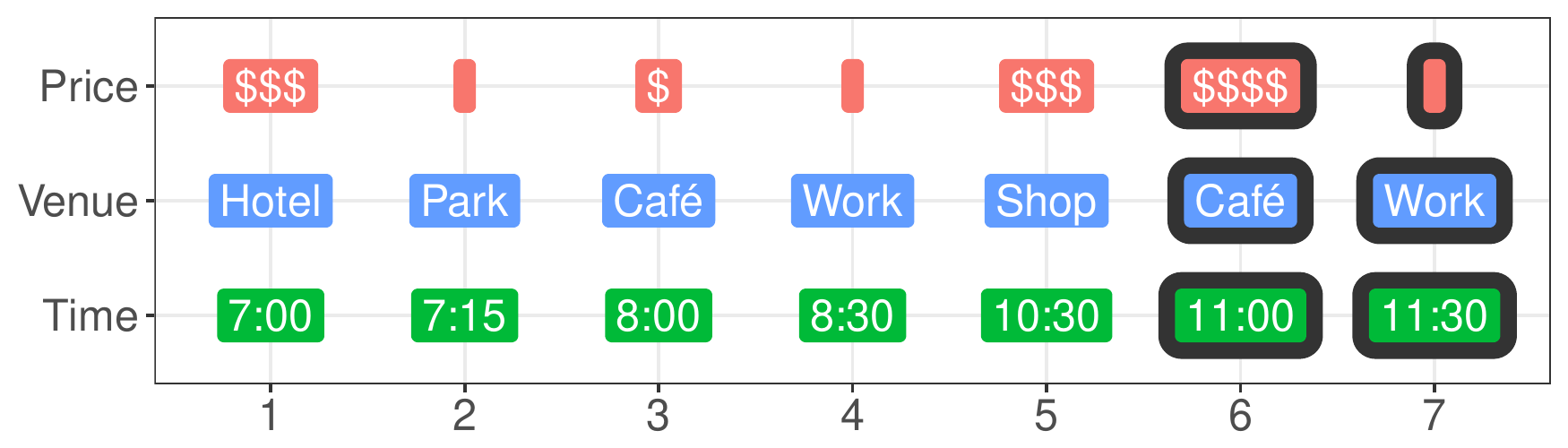}
\caption{Best subtrajectory alignment highlighted into the trajectory.}
\label{fig:bestAlignmentC}
\end{figure}

Considering the previous insights, the question we want to answer is \textit{which of these alignments is the best and how to combine the alignments whether the distances are heterogeneous (time in minutes, venue in number of equal venues and price in units)?} In order to answer this question, avoiding the dimension normalization and weighting, we propose a new \textit{ranking-based approach}. %\emph{This is one of our major contributions}. 
In our approach the best alignment is represented by the check-ins 6 and 7, as highlighted in Figure~\ref{fig:bestAlignmentC}.

In this work we introduce the method \textsc{MasterAlignment} (MUltiple Apect SubTrajEctoRy Alignment), that consists in ranking the distances of each dimension individually and getting the position of the minimum average rank to determine the position of the best alignment. Let us consider $\mathbf{V}_1, \ldots, \mathbf{V}_6$ as the distance vectors of the possible alignments, where $\mathbb{V}_i$ corresponds to the alignment between $s$ and the subtrajectory $r_i$ in $T$ starting at the $i$th-position. Table~\ref{tab:bestalignmentA} presents the distance values of these vectors, where the value at row $Time$ and column $1$ is the distance between $s$ and $r_i$ on dimension $Time$. This value is $435$ because of the sum of difference between time values in minutes, i.e. \textit{(10:30-7:00) = 210} and \textit{(11:00-7:15) = 225}.

\begin{table}[!ht]
\footnotesize
\setlength{\tabcolsep}{4.0pt}
% The Movelet
\centering
\caption{Finding the best alignments from the distance vectors.}
\begin{subtable}[t]{0.48\textwidth}
\caption{Distance values.}
\begin{tabular}{r|p{0.5cm}p{0.5cm}p{0.5cm}p{0.5cm}p{0.5cm}p{0.5cm}}
\hline
& \multicolumn{6}{|c}{Start position alignment} \\
\hline 
Distance & 1 & 2 & 3 & 4 & 5 & 6 \\
\hline
		$Time$ & 435 & 375 & 300 & 150 & 0 & 60 \\
        $Venue$ & 2 & 2 & 0 & 2 & 2 & 0 \\
        $Price$ &  1 &  3 &  3 &  5 & 3 & 2 \\
\hline
Vector & $\mathbf{V}_1$ & $\mathbf{V}_2$ & $\mathbf{V}_3$ & $\mathbf{V}_4$ & $\mathbf{V}_5$ & $\mathbf{V}_6$ \\
\end{tabular}
\label{tab:bestalignmentA}
\end{subtable}
~
\begin{subtable}[t]{0.48\textwidth}
\caption{Distance rankings.}
\begin{tabular}{r|p{0.5cm}p{0.5cm}p{0.5cm}p{0.5cm}p{0.5cm}p{0.5cm}}
\hline
& \multicolumn{6}{|c}{Start position alignment} \\
\hline
Ranking & 1 & 2 & 3 & 4 & 5 & 6 \\
\hline
		$Time$  & 6.0 & 5.0 & 4.0 & 3.0 & 1.0 & 2.0 \\
        $Venue$ & 4.5 & 4.5 & 1.5 & 4.5 & 4.5 & 1.5 \\
        $Price$ & 1.0 & 4.0 & 4.0 & 6.0 & 4.0 & 2.0 \\
\hline 
Avg. rank       & 3.8 & 4.7 & 2.8 & 4.5 & 3.3 & \textbf{1.8} \\
\end{tabular}
\label{tab:bestalignmentB}
\end{subtable}
\label{tab:bestalignment}
\end{table}

Table~\ref{tab:bestalignmentB} shows the distance ranking for each dimension individually, where the value at row $Time$ and column $i$ is the ranking of the distance value at starting position $i$ on dimension $Time$ (among the other alignments at the same dimension). The best ranking at dimension $Time$ is $1.0$ and corresponds to the $5$th starting position, because and dimension $Time$ it has the lowest distance value for dimension $Time$, zero minutes. The second best ranking is $2.0$ and corresponds to the $6$th starting position, because it is the second lowest distance, $60$ minutes. Note that it allows fractional ranks in case of tie, such as $1.5$ at starting positions $3$ and $6$ at dimension $Venue$. After that, it calculates the average rank at each starting position and chooses the lowest average rank as the \emph{best overall ranking}. In this case, at the starting position $6$ it reaches $1.8$ (in bold). Finally, it defines the \emph{distance vector} of the alignment starting at position $6$, which is $W_6=\langle 60, 0, 2\rangle$. 

%Considering that ranking each dimension takes $O(m~log~m)$, where $m$ is the trajectory length. So the complexity to find the best alignment from the \emph{distance vectors} is $O(d \times m~log~m)$, where $d$ is the number of dimensions.

\subsection{Relevance Measuring for Multidimensional Subtrajectory Candidates}
\label{sec:MQM}

%Evaluate the quality of a subtrajectory candidate consists of analyzing its distances to trajectories 
%of the same class and of the opposite class
%in an \emph{orderline} in order to finding the \emph{split point} distance that maximizes the quality score. In the literature have been proposed several techniques to do that, such as maximum information gain~\cite{Ye2011}, Kruskal-Wallis and Mood’s Median~\cite{Lines2012alternative}, and the Left Side Pure~\cite{ferrero2018}. However, all of them are limited to find the split point in a one-dimensional \emph{orderline}.

\begin{figure}[!hbtp]
\footnotesize
\centering
%  trim={<left> <lower> <right> <upper>}
%\includegraphics[width=0.9\textwidth,trim={0cm 0cm 0cm 0cm},clip]{figures/multipleOrderlines.pdf}
\includegraphics[width=0.9\textwidth,trim={0cm 0.0cm 0cm 0.0cm},clip]{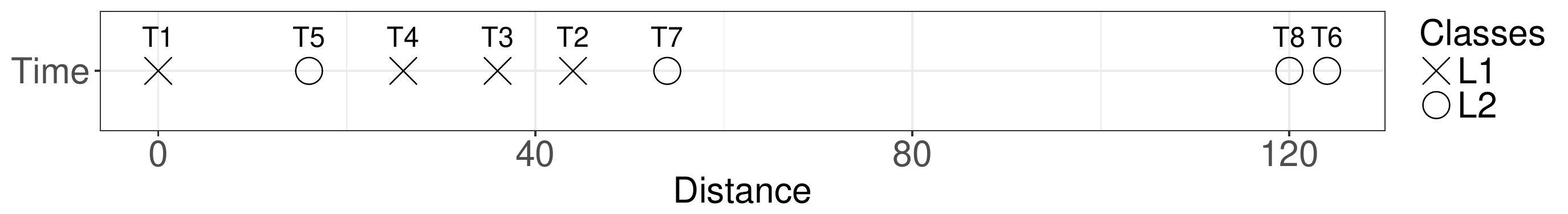}
% dimen~oes do grafico 5.5 x 14
\caption{A one-dimensional orderline.}
\label{fig:orderlineOneDimensional}
\end{figure}

The relevance of a subtrajectory is related to the number of trajectories of the same class that performs similar movement. To define which similar the movement needs to be we need to analyze the distances of the best alignments between a subtrajectory and all trajectories, denoting by $\mathbb{W}$. The most common approach consist of putting the distance in an \emph{orderline} and finding a \emph{split point} to separate the distances into two groups: the nearest and the farthest. Let us consider trajectories $T_1, T_2, \dots T_8$ of classes $L_1, L_2$, represented by a dimension, $Time$, and a subtrajectory $s$ extracted from $T_1$. The trajectories $T_1$ to $T_4$ are of class $L_1$ and the trajectories $T_5$ to $T_8$ are of class $L_2$, and $s$ has $L_1$ as the \emph{target} class, as well as $T_1$. Figure~\ref{fig:orderlineOneDimensional} shows an example of an orderline.

To find the \emph{split point} and measure the relevance in a one dimensional orderline (as in Figure~\ref{fig:orderlineOneDimensional}) have been proposed several techniques, such as the maximum information gain~\cite{Ye2011}, the Kruskal-Wallis and Mood’s Median~\cite{Lines2012alternative}, and the Left Side Pure (LSP)~\cite{ferrero2018}. The LSP returns a \emph{split point} between $T_1$ and $T_5 $ in order to keep the left side of the orderline pure. %Because of the left side only contains $T_1$ the discrimination power of $s$ is low.

\begin{figure}[!hbtp]
\footnotesize
\centering
%  trim={<left> <lower> <right> <upper>}
%\includegraphics[width=0.9\textwidth,trim={0cm 0cm 0cm 0cm},clip]{figures/multipleOrderlines.pdf}
\includegraphics[width=0.9\textwidth,trim={0cm 0cm 0cm 0.0cm},clip]{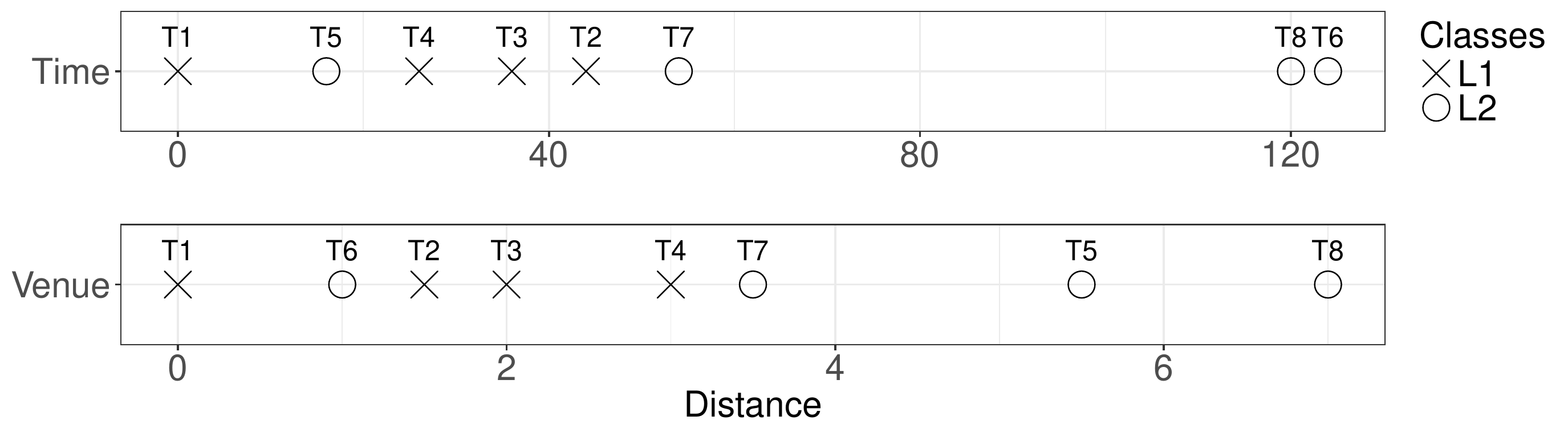}
% dimen~oes do grafico 5.5 x 14
\caption{Multiple one-dimensional orderlines.}
\label{fig:orderlineA}
\end{figure}

Finding the split points for multidimensional subtrajectories is a more complex task, due to the best alignment consist of a set of distances, as proposed in Section~\ref{sec:multidimensionalAlignment}. Let us consider trajectories $T_1, T_2, \ldots T_8$ represented by dimensions $Time$ and $Venue$, and the subtrajectory $s$ extracted from $T_1$. Figure~\ref{fig:orderlineA} shows the representation of the \emph{distance vectors} by an orderline for each dimension. Note that the trajectory $T_1$ is the first distance value on each dimension because $s$ comes from $T_1$.
%The \emph{distance vector} of the best alignment between $s$ and the $j$th subtrajectory is denoted by $W_j$.

One way to analyze both the orderlines is by transforming each \emph{distance vector} in a unique value by normalizing and weighting its dimensions, and finding the best \emph{split point} using the proposed techniques for one-dimensional orderlines~\cite{Ye2011,Lines2012alternative,ferrero2018}. However, this is not a good solution when leading with heterogeneous distances, because of the difficult to define these weights. Another way is to analyze each dimension independently and then find a \emph{split point} for each dimension. But this approach does not consider the interaction between dimensions, which is very important to find \emph{movelets}, and tends to return lower values of \emph{split points}. For instance, in Figure~\ref{fig:orderlineA} both the $Time$ and $Venue$ orderlines have distance values of $L_2$ on the right of $T_1$ distance, which corresponds to trajectories $T_5$ and $T_6$, respectively. By analyzing each orderline independently is difficult to find good \emph{split points}, because it results in a set of \emph{split points} that only separates $T_1$ on the \textit{left side}, which means that the subtrajectory candidate has very low relevance.

\begin{figure}[!htpb]
\centering
\includegraphics[width=1\textwidth]{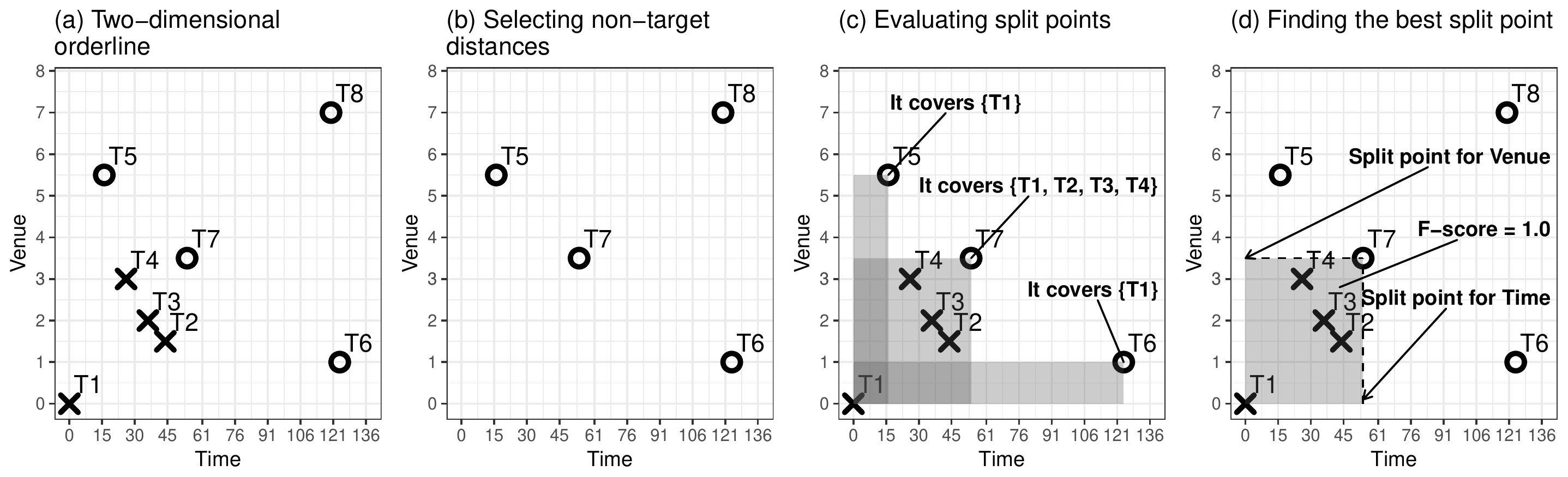}
\caption{Example of finding split points in a \emph{multidimensional orderline}.}
\label{fig:findingSplitpoints}
\end{figure}

Instead of analyzing each orderline separately, we propose a method to analyze all dimensions together. Note in Figure~\ref{fig:orderlineA} the distance vector for $T_7$ may be a good estimator of the split points for $s$, because all the points of class $L_1$ have smaller values. Our method leads with the problem of distances on multiple and heterogeneous dimensions and finds the \emph{split points} that maximizes the relevance of the subtrajectory. The method, called \textsc{MasterRelevance} (MUltilple Aspect SubTrajectoERy Relevance), consist in three steps and it is exemplified by Figure~\ref{fig:findingSplitpoints} using two dimensions $Time$ and $Venue$. 

%We also propose a method to automatically find the split points for all dimensions. 
In Figure \ref{fig:findingSplitpoints}(a) each point in the scatter plot represents the distance vector of a best alignment considering two dimensions, $Time$ and $Venue$, which allows to visualize the interaction between dimensions. The first step consists in selecting only the points of the opposite class $L_2$ and then pruning the points with greater values than some other in both dimensions $Time$ and $Venue$. In Figure~\ref{fig:findingSplitpoints}(b) the point of $T_8$ is pruned because it has greater distance values than $T_7$ for both dimensions. In the second step, it evaluates the unpruned points according the capability to cover points of the \emph{target class} $L_1$ in all dimensions. Figure~\ref{fig:findingSplitpoints}(c) demonstrates that by using the points values of $T_5$ or $T_6$ as the \emph{split points} they only cover $T_1$ of class $L_1$, but by using the point values of $T_7$ they cover $T_1, T_2, T_3,$ and $T_4$. In the final step it chooses the \emph{split points} that cover most points of the \emph{target class} and it calculates the relevance score. As shown in Figure~\ref{fig:findingSplitpoints}(d), the best split points are the values of the point $T_7$ and the score is $1$. To calculate the score we use F-measure that is the harmonic average of the \textit{precision} and the \textit{recall}. In this context, the \textit{precision} is the proportion of \emph{point} covered by the split points that belongs to the \emph{target class} of all covered, and the \textit{recall} is the proportion of \emph{point} covered by the split points that belongs to the \emph{target class} of all \emph{point} of \emph{target class}.

%% file: 4_Experiments.tex
\section{Preliminary Experiments}
\label{sec:experiments}

%We evaluate the proposed approach with two datasets and compare it with five methods in the literature. Section~\ref{sec:datasets} presents the datasets, Section~\ref{sec:experimentalConfiguration} the experimental configuration, and the Section~\ref{sec:results} the results and discussion.

\subsection {Dataset}
\label{sec:datasets}

The Gowalla dataset is a   location-based social networking, where users shared their locations by checking-in~\cite{cho2011friendship}. Each check-in contains the anonymized user id, the timestamp, the location (latitude and longitude), and the check-in ID, without any other information about checking-in\footnote{https://snap.stanford.edu/data/loc-gowalla.html}. This dataset was used in a previous recent work~\cite{gao2017identifying} to classify users based on users' check-in id. From the original dataset containing more than 6mi of check-ins, collected between 2009 and 2010, we cleaned the dataset and segmented the users trajectories into weekly trajectories. We selected check-ins with frequency at least 15, weekly trajectories with at least 10 check-ins and users with at least 10 trajectories, resulting in 33,816 weekly trajectories of 1,952 users. Each check-in is represented by the dimensions: space, time, weekday, and check-in ID. Table~\ref{tab:gowallaDescription} shows details about each dimension, such as data type, values range, and the distance measure used to compare two dimension values. 

\begin{table}[!ht]
\centering
\caption{Gowalla trajectory dimension description.}
\label{tab:gowallaDescription}
\footnotesize
\setlength{\tabcolsep}{5.0pt}
\setlength{\extrarowheight}{3pt}
\vspace{-0.7em}
\begin{tabular}{lllll}
\hline \hline
\textbf{Dimension} & \textbf{Type} & \textbf{Range or examples} & \textbf{Distance measure} \\ 
\hline \hline

Space	& Composite ($lat~lon$) & 40.82651 -73.95039 & Euclidean Distance 
\\ \hline

Time	& Temporal (HH:MM) & [00:00,23:59] & Difference in minutes 
\\ \hline

Weekday & Ordinal & $\{$Mon, Tue, \dots, Sun$\}$ & Weekday Distance\footnote{The weekday distance between two values returns 0 if both are weekdays or weekends, and 1, otherwise.} 
\\ \hline

\makecell[l]{Check-in ID} & Nominal & Any nominal value & Binary Distance \\
\hline \hline
\end{tabular}
\end{table}

\subsection{Experimental Configuration}
\label{sec:experimentalConfiguration}
%Based on the the presented dataset this paper leads with a classification problem of multiple classes (users) based on mobility data represented by multiple and heterogeneous dimensions. There are no works in the literature for leading with this problem. The work proposed in~\cite{gao2017identifying} classify users' trajectories using only the check-in ID information, and does not support to lead with other dimensions.

We compared \textsc{MasterMovelets} with the classifier Bi-TULER proposed in~\cite{gao2017identifying} and classifiers based on four trajectory distance measures: LCSS, EDR, MSM, and DTW. The code of all methods used in this work and the datasets are publicity available in the author's website.

For distance measures we define three values of threshold for each non-nominal dimension, keeping all dimensions with the same weights, and we use the mean and standard deviation of classification accuracy in order to compare them with our method. For evaluating \textsc{MasterMovelets} we used Neural Networks (NN) and Random Forests (RF). The former is a Multilayer Perceptron with a hidden layer by 100 units and to train it we used the same parameters used in~\cite{gao2017identifying}, a dropout rate $0.5$ and an Adam optimizer with the following learning rates $10^{-4}$ (epochs): $9.5(80)$, $7.5(50)$, $5.5(50)$, $2.5(30)$, and $1.5(20)$. The latter consist of an ensemble of 100 decision trees. We evaluate methods on a stratified holdout evaluation separating 70\% for training and 30\% percent to test. For \textsc{MasterMovelets} we only used the training set for movelets discovering. For BiTULER we used the entire dataset to build the word embeddings, as suggested in~\cite{gao2017identifying}.

%For a fair comparison with Bi-TULER we use the code available for authors to build the Recurrent Neural Network and we build embeddings based on the check-ins ID for Gowalla and the Venue Category for Foursquare, using the datasets before selected users for experiments. For the distance measures, we define only two possible values of weights and thresholds for each dimension, totaling 196 runs for LCSS, EDR, and MSM, and 14 for DTW. And we use the values of mean and standard deviation of classification accuracy in order to compare them with our method.

\subsection{Results and Discussion}
\label{sec:results}

Table~\ref{tab:results} shows the experimental results in terms of classification accuracy (\textit{acc}) and accuracy on top 5 (\textit{acc top5}) on the test set. The best result for each dataset and measure are highlighted in bold and the second are underlined.

\begin{table}[!ht]
\footnotesize
\setlength{\tabcolsep}{3.5pt}
\setlength{\extrarowheight}{4pt}
\caption{Experimental results }
\label{tab:results}
\centering
\vspace{-1em}
\begin{tabular}{lc|ccccc|cc}
  \hline \hline
  & &  &  &  &  &  & \multicolumn{2}{c}{\makecell[c]{\textsc{Master}\\ \textsc{Movelets}}} \\
  Dataset & Measure & DTW & LCSS & EDR & MSM & BiTULER & NN & RF \\  
  \hline \hline

\multirow{2}{*}{Gowalla}
& \textit{acc} 		& 73.7(0.8) & 87.0(2.5) & 80.9(3.9) & 89.8(1.1) & 34.7 & \textbf{95.2} & \underline{92.6} \\
& \textit{acc top5} & 86.7(0.8) & 91.0(2.5) & 83.9(3.9) & 93.6(1.1) & 56.0 & \textbf{98.2} & \underline{97.8} \\
\hline
%\multirow{2}{*}{Foursquare}
%& \textit{acc} 		& 15.4(1.1) & 22.4(1.6) & 25.0(0.9) & 40.5(1.6) & 9.3 & \textbf{77.9} & \underline{66.9} \\
%& \textit{acc top5} & 30.0(1.6) & 44.3(2.0)	& 45.1(1.4) & %63.0(1.1)	& 21.8 & \textbf{89.7} & \underline{84.9} \\ 
%\hline \hline
\end{tabular}
\end{table}

The results show that the use of \textsc{MasterMovelets} to build NN and RF models outperforms the state of the art methods on both datasets. BiTULER presents the worst classification results in both datasets. These classification problems involve multiple dimensions and BiTULER is limited to only consider one, the checkin ID information. DTW uses all dimensions, but it presents the worst results among the distance measures, because it is difficult to weight the distances of all multiple and heterogeneous dimensions. The distance measures LCSS and EDR also do not present good classification accuracy, since the chance of have matching decreases as the number of dimensions increases. The MSM present better results than the LCSS and EDR, because it allows partial matching among dimensions, so it is less affected than those. Both of models built from \textsc{MasterMovelets} present the best results, however the Neural Network model better capture the relation between the \emph{movelets} and the class. Our NN model only has a hidden layer with 100 units, which is much simpler than the Bidirectional LSTM Recurrent Neural Network with 300 units in the hidden layer, proposed in~\cite{gao2017identifying}. %Furthermore, the training process of our neural networks models converges in the first 80 iterations. 
In general, Neural Networks models lead better with high dimensional spaces than symbolic models, in detriment of its interpretability. RF also achieve better results than the state-of-the-art methods. This model consists of ensembles of decision trees, that allow to extract relevant classification rules.

%% file: 5_Conclusions.tex
\vspace{-1em}
\section{Conclusions}
\label{sec:conclusions}

\begin{sloppypar}
In this paper we proposed a new method for extracting relevant subtrajectories for multiple aspect trajectory classification, called \textsc{MasterMovelets}. The method consists of an extension of a previous work to extract relevant subtrajectories from raw trajectories. Our method finds the most relevant subtrajectories leading with the problem of exploring dimension combinations. \textsc{MasterMovelets} is parameter-free and domain independent, which is very important since parameters values are difficult to estimate in many problems and directly affect the data mining results.

We evaluate our method on a dataset of check-ins  to classify user's trajectories and compare it with several methods in the literature for trajectory classification. Initial results demonstrate that \emph{MasterMovelets} is very promising. This is the first work in the literature for classifying trajectories represented by multiple and heterogeneous dimensions, recently introduced as Multiple Aspects Trajectories.

Future works include a more robust experimental evaluation and improvement of the  complexity analysis of movelets discovery, that is the main drawback of our method.

\end{sloppypar}